%% file: main_arxiv.tex
\documentclass[letterpaper]{article} 
\usepackage[preprint]{aaai2027}  
\usepackage[hyphens]{url}  
\usepackage{graphicx} 
\usepackage{natbib}  
\usepackage{caption} 
\usepackage{amsmath}
\usepackage{amssymb}
\usepackage{booktabs}      
\usepackage{multirow}
\usepackage{algorithm}
\usepackage{algpseudocode}

\usepackage{newfloat}
\usepackage{listings}
\DeclareCaptionStyle{ruled}{labelfont=normalfont,labelsep=colon,strut=off} 

\floatstyle{ruled}
\newfloat{listing}{tb}{lst}{}
\floatname{listing}{Listing}

\usepackage{booktabs}

\title{CausalGate: Causal Importance Distillation for Transformer Module Pruning}

\author{
    Kiran Nair\corresponding,
    Smriti Regmi,
    Rodrigue Rizk
}

\affiliations{
    USD Artificial Intelligence Research Lab, Department of Computer Science\\
    University of South Dakota, Vermillion, SD 57069, USA\\
    kiran.prasannannair@coyotes.usd.edu
}

\begin{document}

\maketitle

\begin{abstract}
Existing adaptive inference methods for Large Language Models rely on observational heuristics, such as hidden-state similarity or activation magnitudes, to drop redundant modules. However, these correlation-based metrics often fail to capture subtle, non-linear structural computations vital for semantic accuracy. We introduce CausalGate, an intervention-guided framework for compute-efficient transformer inference. During a calibration phase, CausalGate isolates individual Attention and MLP sub-layers, zeros out their respective outputs, and measures the exact semantic damage via the Kullback-Leibler divergence of the final logit distribution. To eliminate runtime routing overhead, this structural importance hierarchy is distilled into a global set of static, lightweight scalar gates using an Exponential Moving Average smoothing objective paired with a differentiable pairwise ranking loss. Evaluated on TinyLlama-1.1B, Qwen2.5-3B, and Llama-3.1-8B across language modeling and commonsense reasoning benchmarks, CausalGate consistently outperforms prominent dynamic routing and layer-skipping baselines, translating theoretical compute savings into concrete hardware latency reductions with zero operational overhead.
\end{abstract}

\section{Introduction}
Ever since the seminal paradigm shift of "Attention Is All You Need,"~\cite{vaswani2017attention} scaling transformer architectures has unlocked unprecedented linguistic capabilities at a severe computational cost, making Large language model (LLM) inference a critical deployment bottleneck. To mitigate this, model compression methods such as structural pruning and sub-layer skipping~\cite{he2025adaskip} seek to permanently deactivate low-utility blocks, avoiding the heavy hardware execution overhead and branch prediction penalties characteristic of runtime-dynamic routing frameworks like CALM~\cite{schuster2022confident} or Mixture-of-Depths (MoD)~\cite{raposo2024mixture}. Existing structural pruning metrics typically rely on observational proxies, such as activation magnitudes or hidden-state similarity, to determine if a module can be bypassed. However, such observational metrics can occasionally overlook subtle yet critical computational steps. A hidden-state vector traversing a module might exhibit minimal geometric displacement, yet that module may still execute a non-linear "causal refinement" vital for factual accuracy or logical coherence. By relying primarily on representational characteristics rather than direct structural necessity, heuristic-driven static pruning risks bypassing quietly essential modules, leading to sharp performance degradation at higher compute savings.

To overcome the limitations of observational proxies, we look to classical scientific inquiry: to truly determine whether a component is essential to a system, one must actively intervene upon it rather than merely observe it. We introduce CausalGate, an offline structural calibration framework establishing a permanent module-selection hierarchy through direct, input-agnostic structural interventions. Instead of estimating utility from runtime representation alignments, CausalGate isolates individual Attention and MLP modules during an offline calibration phase, zeros out their outputs, and measures the resulting Kullback-Leibler ($KL$) divergence ($D_{\text{KL}}$) in the final token distribution. This approach is governed by a central \textit{\textbf{hypothesis:} a module's true global computational importance is explicitly defined by its causal influence on the final logit distribution.} By operationalizing this principle at the sub-layer level, CausalGate bypasses heuristic guesswork and directly quantifies the exact semantic damage incurred by skipping any given block. 

A key architectural distinction of CausalGate is its fine-grained operation at the sub-layer module level rather than the monolithic transformer-layer level. Because Multi-Head Attention and MLP modules execute distinct functions, attention routes context across sequences while MLPs act as localized Key-Value (KV) memories, independent pruning yields significantly higher computational flexibility than uniform layer-dropping. However, calculating causal interventions at runtime is prohibitive. To achieve zero-overhead inference, CausalGate distills the structural importance hierarchy discovered during calibration into global, static, learned scalar gates. We optimize these gates using an Exponential Moving Average (EMA) smoothing objective to filter intervention noise, paired with a differentiable pairwise ranking loss that forces gate values to strictly preserve the discovered causal ordering.

This optimization strategy bridges causal analysis and hardware execution efficiency, encoding global structural knowledge into a deterministic inference mask. Because these functional sub-layers perform decoupled tasks, the model establishes a sequence-agnostic importance hierarchy operating at compile-time. Consequently, CausalGate eliminates runtime routing overhead, ensuring straight-line inference execution that pairs exceptionally well with localized memory boundaries and standard hardware loops. In summary, our contributions are as follows:
\begin{enumerate}
    \item We introduce a novel approach to structural model pruning that replaces traditional observational heuristics with a ground-truth importance hierarchy derived directly from structural causal interventions during calibration.
    \item We demonstrate that evaluating and skipping Attention and MLP modules independently yields a significantly more flexible and precise compute-performance tradeoff than uniform layer-dropping strategies.
    \item We propose an optimization framework utilizing EMA target smoothing and a pairwise ranking loss to successfully encode structural causal hierarchies into global, static gates, enabling zero-overhead inference.
    \item We conduct extensive evaluations against prominent baselines, demonstrating that CausalGate preserves language-model quality substantially better at larger compute reductions.
\end{enumerate}
The remainder of this paper is organized as follows. Section~\ref{sec:related_work} reviews related literature in adaptive inference and mechanistic interpretability. Section~\ref{sec:methodology} introduces the structural components of the CausalGate architecture alongside its intervention and training frameworks. Section~\ref{sec:experiments} presents our empirical evaluation and comparative baseline analyses. Finally, the broader implications, limitations, and avenues for future work are discussed in Section~\ref{sec:discussion}, and Section~\ref{sec:conclusion} provides concluding summaries.

\section{Related Works}
\label{sec:related_work}
\paragraph{Adaptive Inference and Layer Skipping.}
Adaptive inference dynamically adjusts computation based on input difficulty or latency bounds. Early mechanisms in Convolutional Neural Networks leveraged filter pruning, quantization~\cite{zhang2024appq}, early exiting~\cite{han2021dynamic,bolukbasi2017adaptive}, and dynamic channel gating~\cite{bejnordi2019batch,gao2018dynamic,lin2017runtime}. In Vision Transformers, frameworks like DynamicViT~\cite{rao2021dynamicvit} and A-ViT~\cite{yin2022vit} focus on dynamic token reduction, while AdaViT~\cite{meng2022adavit} adaptively drops attention heads or entire blocks per sample. For Large Language Models (LLMs), dynamic computational routing has expanded into token-skipping policies such as Mixture-of-Depths (MoD)~\cite{raposo2024mixture} and conditional early-exiting schemes~\cite{rotem2023finding}. Closer to our work, structural alternatives focus on bypassing redundant segments via uniform layer skipping~\cite{liu2024accelerating,luo2025adaptive}, input-output similarity heuristics~\cite{he2025adaskip}, reinforcement learning routing~\cite{liu2025aster}, or training-infused residual routing gates~\cite{laitenberger2025layers,nair2026spike}. However, these approaches predominantly introduce dynamic control-flow branching or rely on shallow observational proxies during runtime. In contrast, we shift the execution boundary entirely to compile-time by utilizing offline interventional distillation to generate zero-overhead, static module masks.

\paragraph{Structural Pruning.}
Static network compression permanently drops underutilized weights or channels. For LLMs, post-training pruning often prioritizes channel groups, layers, or individual weights based on raw magnitudes~\cite{frantar2022gptq}, weight-activation products like Wanda~\cite{sun2024simple}, or structural activation variances. To compress architectures structurally, frameworks like LLM-Pruner~\cite{ma2023llm} and Sheared LLaMA~\cite{xia2024sheared} utilize first-order gradient approximations to isolate redundant coupled components for removal. Similarly, static sensitivity analysis leverages Taylor-series expansions or zero-shot similarity metrics, such as Block Influence in ShortGPT~\cite{men2025shortgpt}, to map module importance. While related, these alternatives rely strictly on observational correlations or localized gradients. CausalGate fundamentally differs by operationalizing active interventional calculus ($do$-operations) at a decoupled Attention and MLP sub-layer granularity to isolate true downstream semantic shifts.

\paragraph{Complementary Acceleration Paradigms.}
Beyond layer pruning, the literature contains orthogonal strategies to alleviate LLM inference bottlenecks. Memory footprints are frequently managed via compressed or pyramidal KV caches~\cite{ge2024model,xiao2025duoattention,cai2024pyramidkv}, or via uniform mixed-precision weight and activation quantization frameworks~\cite{xiao2023smoothquant,lin2024awq,frantar2022gptq,park2024lut}. Concurrently, speculative and aggressive decoding paradigms~\cite{leviathan2023fast,stern2018blockwise,sun2021instantaneous} accelerate autoregressive generation by predicting multiple tokens in parallel via distinct draft verification branches. These methodologies remain fully complementary to CausalGate and can be applied jointly to maximize hardware deployment efficiency.

\section{Methodology}
\label{sec:methodology}
Our key intuition is that modules whose removal substantially alters the output distribution should be preserved, whereas modules with negligible influence can be skipped with minimal impact on model quality.
\setlength{\textfloatsep}{8pt plus 2pt minus 2pt}
\setlength{\intextsep}{8pt plus 2pt minus 2pt}
\begin{figure*}[t]
    \centering

    \begin{minipage}[t]{0.18\textwidth}
        \centering
        \includegraphics[width=\linewidth]{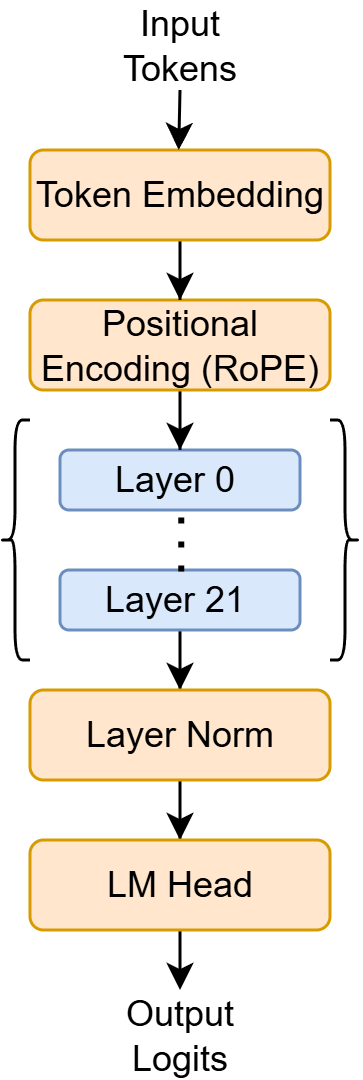}
        \caption{TinyLlama backbone.}
        \label{fig:overview_backbone}
    \end{minipage}
    \hfill
    \begin{minipage}[t]{0.50\textwidth}
        \centering
        \includegraphics[width=\linewidth]{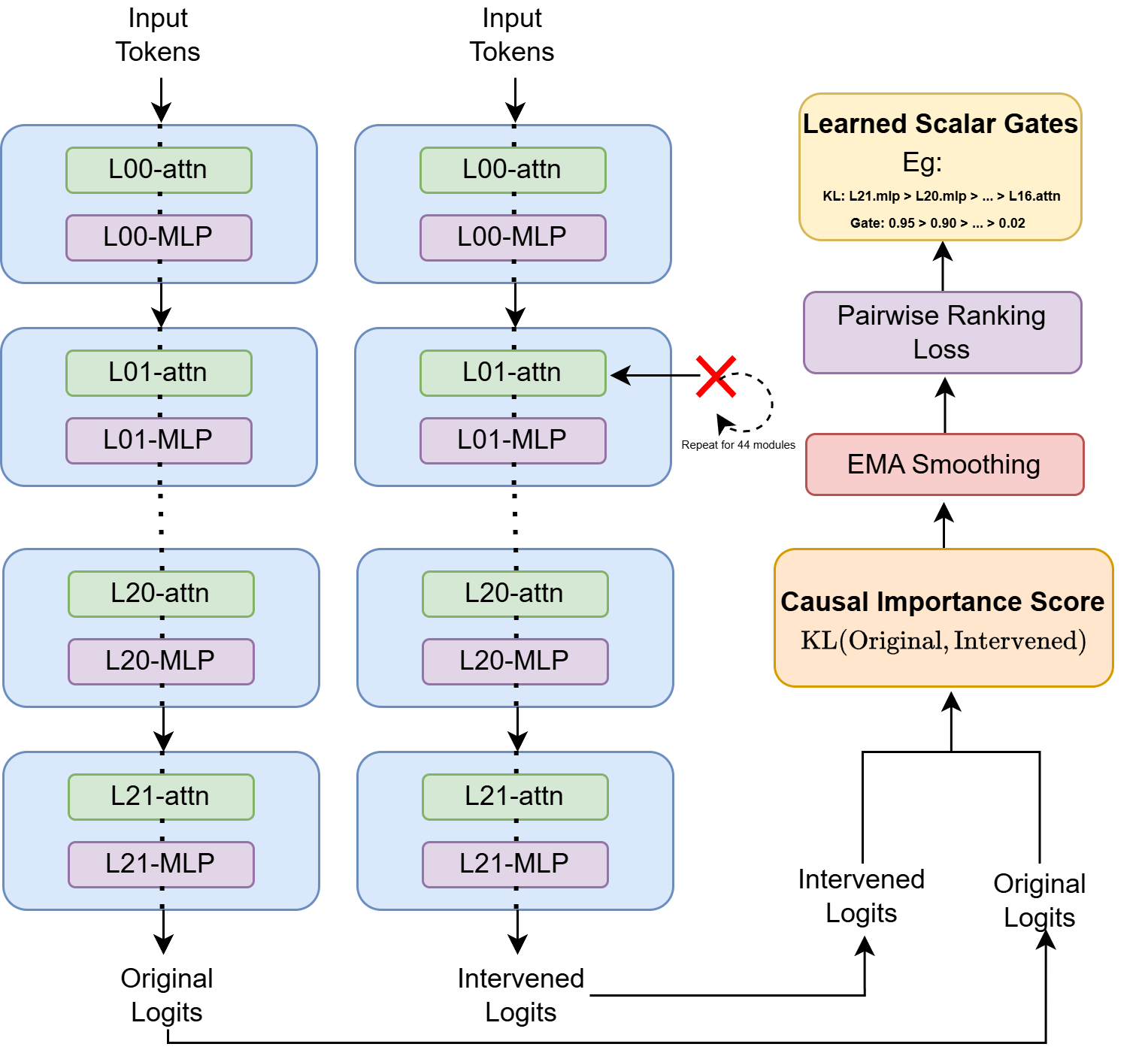}
        \caption{Intervention-based causal importance estimation and gate learning.}
        \label{fig:overview_causal_gate}
    \end{minipage}
    \hfill
    \begin{minipage}[t]{0.22\textwidth}
        \centering
        \includegraphics[width=\linewidth]{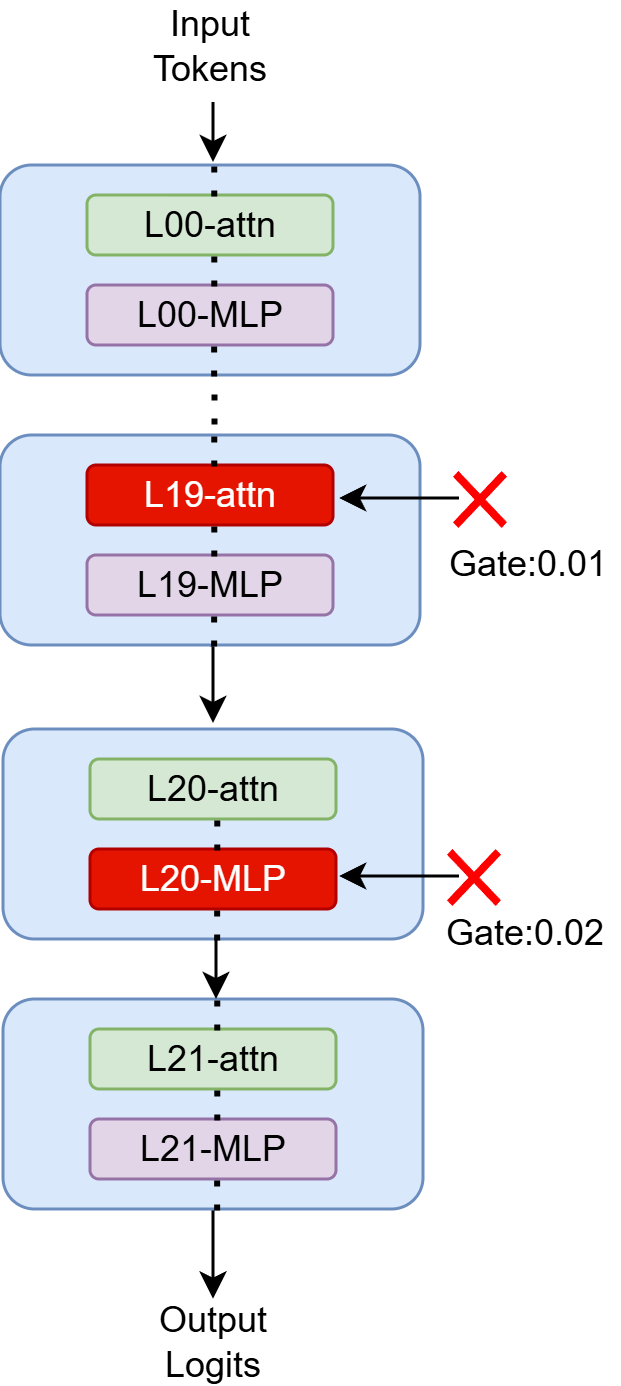}
        \caption{Compute-efficient inference by skipping low-gate modules.}
        \label{fig:overview_inference}
    \end{minipage}

    \caption{
    Proposed intervention-guided module selection framework. Each attention and MLP module is independently evaluated by zeroing its output; the resulting final logit KL divergence defines its causal importance score. EMA smoothing and pairwise ranking losses distill these signals into learned scalar gates, allowing low-utility modules to be skipped at inference under a target compute budget.
    }
    \label{fig:method_overview}
\end{figure*}

Let a transformer model consist of $L$ layers, each containing an attention submodule and an MLP submodule. We treat each attention and MLP block as an independent computational unit, yielding a set of $M$ modules $\mathcal{M}=\{m_1,m_2,\ldots,m_M\}$. For TinyLlama-1.1B, this corresponds to $M=44$ modules (22 attention and 22 MLP modules). Our framework consists of three stages, as illustrated in Figure~\ref{fig:method_overview}: (1) intervention-based causal importance estimation, (2) distillation of causal importance into learned module gates, and (3) compute-efficient inference through module selection.

\subsection{Causal Module Importance Estimation}

We estimate the importance of each module through intervention analysis. Given an input sequence $x$, we first obtain the output distribution of the original model, $p(y \mid x).$ Next, we intervene on a module $m_i$ by zeroing its output activation while leaving all remaining modules unchanged. The resulting model produces an intervened output distribution $p(y \mid x, m_i \rightarrow 0)$. The causal importance of module $m_i$ is quantified as the KL divergence between the intervened and original output distributions:
\begin{equation}
\Delta_i^{(t)}=D_{\mathrm{KL}}\Big(p(y \mid x, m_i \rightarrow 0)\;\|\;p(y \mid x)
\Big),
\label{eq:causal_importance}
\end{equation}
where $t$ denotes the current training iteration. A large value of $\Delta_i^{(t)}$ indicates that intervening on module $m_i$ substantially alters the model's predictive distribution, suggesting that the module plays an important role in generation. Conversely, a small value implies that the module has relatively little influence on the final prediction. We compute intervention scores independently for every attention and MLP module in the network. The resulting set of scores forms a module-level causal importance map that characterizes the relative contribution of each transformer module to the model's predictive behavior.

\subsection{Causal Importance Distillation}
Although intervention scores provide a principled estimate of module importance, they can exhibit substantial variability across training iterations. Consequently, directly using raw intervention scores for module selection can lead to unstable rankings and suboptimal compute-allocation decisions. To address this issue, we distill intervention-derived causal importance into a set of learned module gates that capture a stable global ordering of module relevance. For each module $m_i$, we associate a learnable scalar gate
\begin{equation}
g_i = \sigma(\theta_i),
\end{equation}
where $\theta_i$ is a trainable logit and $\sigma(\cdot)$ denotes the sigmoid function. Each gate is a single learned scalar associated with a transformer module and is shared across all inputs and token positions. Higher gate values indicate modules that should be preferentially preserved during inference.

\paragraph{EMA Target Aggregation.}
Intervention-derived importance estimates can vary across training iterations, resulting in unstable module rankings. To obtain more reliable supervision, we first normalize the intervention scores using the maximum module importance in the current iteration:
\begin{equation}
\tilde{z}_i^{(t)}=\frac{\Delta_i^{(t)}}{\max_j \Delta_j^{(t)} + \epsilon},
\label{eq:max_norm}
\end{equation}
where $\epsilon$ is a small constant for numerical stability. A target floor is then applied to prevent low-importance modules from collapsing to zero:
\begin{equation}
z_i^{(t)}=\alpha+(1-\alpha)\tilde{z}_i^{(t)},
\label{eq:target_floor}
\end{equation}
where $\alpha$ denotes the target floor. The final target is updated using an EMA:
\begin{equation}
\bar{z}_i^{(t)}=\beta \bar{z}_i^{(t-1)}+(1-\beta) z_i^{(t)},
\label{eq:ema}
\end{equation}
where $\beta \in [0,1)$ is the EMA decay factor.  The EMA targets are initialized as $\bar{z}_i^{(0)} = z_i^{(0)}$. The learned gates are then trained to match the EMA-smoothed targets using a mean-squared error objective:
\begin{equation}
\mathcal{L}_{\mathrm{MSE}}=\frac{1}{M}\sum_{i=1}^{M}
\left(g_i-\bar{z}_i^{(t)}\right)^2,
\label{eq:mse_loss}
\end{equation}
where $M$ denotes the total number of modules.

\paragraph{Pairwise Ranking Supervision.}
While the absolute magnitude of intervention scores may vary, the relative ordering between modules is often more informative for module selection. We therefore introduce a pairwise ranking objective that encourages the learned gates to preserve the causal ordering induced by the EMA-smoothed targets. For a sampled module pair $(m_i,m_j)$, we define
\begin{equation}
s_{ij}=\mathrm{sign}\!\left(\bar{z}_i^{(t)}-\bar{z}_j^{(t)}\right),
\end{equation}
where tied pairs are ignored. The pairwise ranking loss is then
\begin{equation}
\mathcal{L}_{\mathrm{rank}}=\frac{1}{|\mathcal{P}|}
\sum_{(i,j)\in\mathcal{P}}\max\left(0,\,\gamma-s_{ij}(g_i-g_j)\right),
\label{eq:ranking_loss}
\end{equation}
where $\mathcal{P}$ is the set of sampled non-tied module pairs and $\gamma$ is the ranking margin. This objective penalizes pairs whose gate ordering is reversed or whose correctly ordered gate values are separated by less than the margin $\gamma$. The sparsity regularizer encourages smaller gate activations and is defined as
\begin{equation}
\mathcal{L}_{\mathrm{sparse}}=\frac{1}{M}\sum_{i=1}^{M}g_i^2.
\label{eq:sparse_loss}
\end{equation}
The final gate-training objective combines language modeling, sparsity, causal regression, and ranking supervision:
\begin{equation}
\mathcal{L}=\lambda_{\mathrm{lm}}\mathcal{L}_{\mathrm{LM}}
+\lambda_{\mathrm{sparse}}\mathcal{L}_{\mathrm{sparse}}
+\lambda_{\mathrm{causal}}\mathcal{L}_{\mathrm{MSE}}
+\lambda_{\mathrm{rank}}\mathcal{L}_{\mathrm{rank}}.
\label{eq:total_loss}
\end{equation}

\begin{figure}[t!]
\centering
\includegraphics[width=\columnwidth]{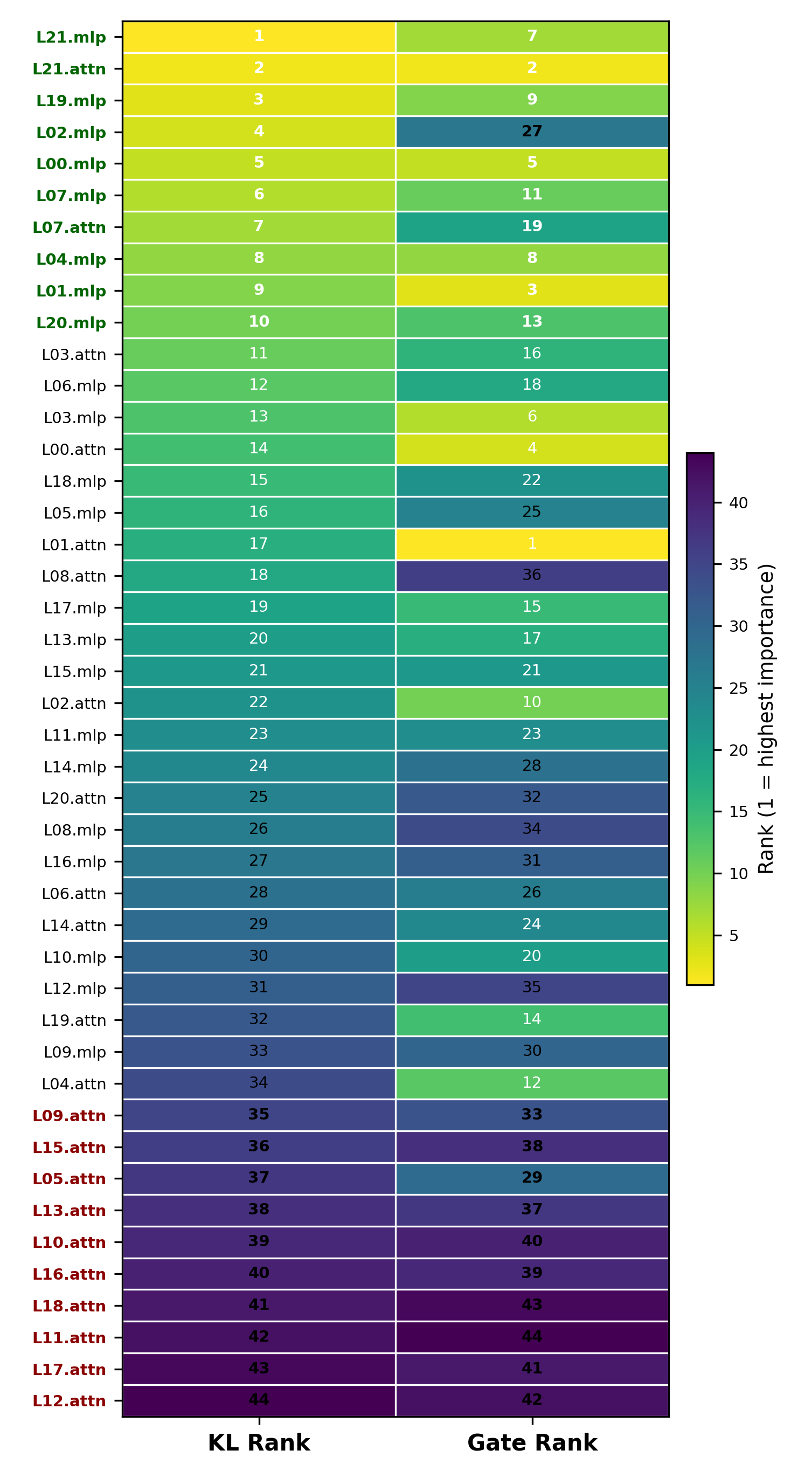}
\caption{
Comparison between intervention-derived causal importance rankings and the rankings learned by CausalGate.}
\label{fig:gate_rank_heatmap}
\end{figure}
\subsection{CausalGate Inference}
After training, the learned gate values provide a global ranking of module importance that can be directly leveraged for compute-efficient inference. Let $\mathcal{G}
=\{g_1,g_2,\ldots,g_M\}$ denote the set of learned gate values for the $M$ transformer modules. Modules are sorted according to their gate values, producing an ordered importance ranking $g_{(1)} \geqslant g_{(2)} \geqslant \cdots \geqslant g_{(M)}$, where larger gate values correspond to modules with higher estimated causal importance. Given a target module-skip ratio, we retain only the highest-ranked modules and disable the remaining modules. Specifically, for a target compute reduction ratio $\rho$, the number of modules to retain is
\begin{equation}
K = \left\lceil (1-\rho)M \right\rceil.
\end{equation}
The top-$K$ modules according to the learned ranking are preserved, while the remaining modules are skipped during inference. Formally, the binary inference mask for module $m_i$ is defined as
\begin{equation}
b_i =
\begin{cases}
1, & \text{if } g_i \in \operatorname{TopK}(\mathcal{G},K), \\
0, & \text{otherwise}.
\end{cases}
\label{eq:binary_mask}
\end{equation}
where $b_i=1$ indicates that the module remains active and $b_i=0$ indicates that the module is skipped.

\begin{table*}[t]
\centering
\small
\caption{Performance comparison across progressive module-removal compute budgets on TinyLlama-1.1B. Lower perplexity is better ($\downarrow$), while higher accuracy is better ($\uparrow$). All baselines are evaluated at matched architectural savings tiers.}
\label{tab:main_results_all}
\begin{tabular}{clcccccc}
\toprule
\textbf{Budget} & \textbf{Method} & \textbf{WikiText PPL}$\downarrow$ & \textbf{C4 PPL}$\downarrow$ & \textbf{HellaSwag}$\uparrow$ & \textbf{PIQA}$\uparrow$ & \textbf{CSQA}$\uparrow$ & \textbf{WinoGrande}$\uparrow$ \\ \midrule
0\% & TinyLlama-1.1B (Ref) & 13.58 & 9.89 & 0.484 & 0.707 & 0.367 & 0.540 \\
\midrule
\multirow{6}{*}{\textbf{5\%}} 
 & \textbf{CausalGate (Ours)} & 16.00 & \textbf{10.84} & 0.445 & 0.723 & \textbf{0.359} & \textbf{0.54} \\
 & CALM (Softmax) & 32.33 & 20.85 & 0.383 & 0.664 & 0.320 & 0.524 \\
 & CALM (Hidden-State) & 23.11 & 28.97 & 0.453 & \textbf{0.734} & 0.313 & 0.540 \\
 & MoD & 25.04 & 16.00 & \textbf{0.484} & 0.648 & \textbf{0.359} & 0.516 \\
 & GateSkip & \textbf{14.71} & 16.12 & 0.461 & 0.684 & 0.313 & \textbf{0.54} \\
 & AdaSkip & 27.08 & 26.42 & 0.406 & 0.633 & 0.273 & 0.510 \\
 & Act-Norm (Zero-Shot) & 25.26 & 19.59 & 0.359 & 0.606 & 0.301 & 0.520 \\
\midrule
\multirow{6}{*}{\textbf{10\%}} 
 & \textbf{CausalGate (Ours)} & \textbf{18.46} & \textbf{12.84} & 0.410 & \textbf{0.715} & 0.332 & 0.506 \\
 & CALM (Softmax) & 71.96 & 48.35 & 0.383 & 0.648 & 0.297 & 0.522 \\
 & CALM (Hidden-State) & 46.43 & 70.59 & 0.391 & 0.641 & 0.281 & \textbf{0.538} \\
 & MoD & 60.29 & 40.75 & 0.406 & 0.605 & \textbf{0.336} & 0.492 \\
 & GateSkip & 22.43 & 27.34 & \textbf{0.441} & 0.613 & 0.293 & 0.518 \\
 & AdaSkip & 40.40 & 43.75 & 0.410 & 0.602 & 0.246 & 0.514 \\
 & Act-Norm (Zero-Shot) & 81.11 & 161.86 & 0.262 & 0.512 & 0.234 & 0.490 \\
\midrule
\multirow{6}{*}{\textbf{20\%}} 
 & \textbf{CausalGate (Ours)} & \textbf{67.0} & \textbf{40.0} & 0.340 & \textbf{0.610} & 0.260 & 0.490 \\
 & CALM (Softmax) & 198.0 & 141.0 & 0.350 & \textbf{0.610} & \textbf{0.300} & 0.510 \\
 & CALM (Hidden-State) & 80.0 & 130.0 & \textbf{0.370} & 0.590 & 0.250 & \textbf{0.520} \\
 & MoD & 407.0 & 269.0 & 0.310 & 0.510 & 0.260 & \textbf{0.520} \\
 & GateSkip & 81.0 & 100.0 & 0.320 & 0.550 & 0.260 & 0.480 \\
 & AdaSkip & 1816.0 & 531.0 & 0.246 & 0.554 & 0.200 & 0.490 \\
 & Act-Norm (Zero-Shot) & 2781.6 & 1085.0 & 0.273 & 0.535 & 0.231 & 0.510 \\
\midrule
\multirow{6}{*}{\textbf{30\%}} 
 & \textbf{CausalGate (Ours)} & \textbf{116.0} & \textbf{69.0} & \textbf{0.340} & 0.540 & 0.250 & 0.490 \\
 & CALM (Softmax) & 601.0 & 422.0 & 0.290 & \textbf{0.600} & \textbf{0.290} & 0.500 \\
 & CALM (Hidden-State) & 185.0 & 299.0 & 0.290 & \textbf{0.600} & 0.230 & \textbf{0.520} \\
 & MoD & 1313.0 & 990.0 & 0.250 & 0.530 & 0.240 & 0.510 \\
 & GateSkip & 365.0 & 491.0 & 0.257 & 0.520 & 0.250 & 0.470 \\
 & AdaSkip & 2064.0 & 623.0 & 0.238 & 0.539 & 0.170 & 0.490 \\
 & Act-Norm (Zero-Shot) & 3563.2 & 3385.3 & 0.238 & 0.535 & 0.176 & 0.500 \\
\midrule
\multirow{6}{*}{\textbf{40\%}} 
 & \textbf{CausalGate (Ours)} & \textbf{1365.0} & \textbf{441.0} & \textbf{0.280} & 0.500 & 0.200 & 0.480 \\
 & CALM (Softmax) & 1574.0 & 1047.0 & \textbf{0.280} & 0.530 & \textbf{0.250} & \textbf{0.500} \\
 & CALM (Hidden-State) & 1472.0 & 712.0 & 0.270 & \textbf{0.570} & 0.220 & 0.490 \\
 & MoD & 4433.0 & 3403.0 & 0.260 & 0.510 & 0.180 & \textbf{0.500} \\
 & GateSkip & 1698.0 & 2046.0 & 0.250 & 0.490 & 0.200 & 0.460 \\
 & AdaSkip & 4260.0 & 4509.0 & 0.235 & 0.527 & 0.160 & 0.460 \\
 & Act-Norm (Zero-Shot) & 20825.4 & 31094.0 & 0.254 & 0.527 & 0.207 & 0.488 \\
\bottomrule
\end{tabular}
\end{table*}

\section{Experiments}
\label{sec:experiments}
We evaluate whether intervention-guided module ranking can identify computationally important transformer modules while preserving language modeling and reasoning performance under reduced compute budgets. Specifically, we seek to answer the following questions:
\begin{itemize}
\item Can CausalGate retain model quality while disabling low-importance modules?
\item How does CausalGate compare against existing compute-reduction baselines?
\item Does intervention-derived supervision produce a meaningful ranking of module importance?
\end{itemize}
\subsection{Experimental Setup}
\paragraph{CausalGate Setup.}All experiments are conducted using TinyLlama-1.1B, which consists of 22 transformer layers and 44 computational modules (22 attention and 22 MLP submodules). Gates are trained using AdamW with a learning rate of 0.01, gradient accumulation of 8 steps, and a maximum of 1000 optimization steps. We employ an EMA decay factor of $\beta=0.9$, target floor $\alpha=0.25$, ranking margin $\gamma=0.05$, and 128 sampled ranking pairs per iteration. The loss weights are set to $\lambda_{\mathrm{LM}}=1.0$, $\lambda_{\mathrm{sparse}}=0.001$, $\lambda_{\mathrm{causal}}=10.0$, and $\lambda_{\mathrm{rank}}=2.0$. All experiments, including baseline evaluations, were conducted on a single NVIDIA Tesla V100 GPU with 32\,GB of memory.  The hyperparameter configuration was selected using a one-factor-at-a-time (OFAT) sensitivity analysis, in which each hyperparameter was varied independently while all remaining settings were fixed. The complete sensitivity analysis and additional experimental results are provided in the Appendix (Table~\ref{tab:hp_sensitivity}).

\paragraph{Baselines.}We compare CausalGate against representative compute-reduction strategies spanning adaptive-depth inference, dynamic routing, learned gating, and heuristic module selection. The full TinyLlama-1.1B model serves as the no-compression reference. We include two CALM variants~\cite{schuster2022confident} based on confidence-driven early exiting: (i) Softmax Margin, which exits when prediction confidence exceeds a calibrated threshold, and (ii) Hidden-State Saturation, which exits when consecutive hidden representations become sufficiently similar. To evaluate dynamic depth allocation, we implemented a Mixture-of-Depths (MoD)~\cite{raposo2024mixture} routing baseline that selectively bypasses computations based on token-level routing decisions. We further compare against GateSkip~\cite{laitenberger2025layers}, a learned gating approach that trains lightweight gating functions to suppress low-importance computations, and AdaSkip~\cite{he2025adaskip}, a heuristic module-selection strategy that ranks modules according to input-output  similarity. To further validate the precise structural contribution of our proposed intervention framework over basic observational methods, we also implement a zero-shot Activation-Norm Pruning (Act-Norm) baseline. This approach executes a forward pass over the calibration data to measure the absolute magnitude of the activations flowing out of each attention and MLP module. Formally, let $\mathbf{H}_{i} \in \mathbb{R}^{T \times d}$ represent the output activation matrix of module $m_i$ for a sequence of length $T$. The zero-shot importance metric $I_{\text{zero}}(m_i)$ is computed as the expected magnitude across the token dimension:
\begin{equation}
I_{\text{zero}}(m_i) = \frac{1}{T} \sum_{t=1}^{T} \|\mathbf{H}_{i,t}\|_2^2
\label{eq:act_norm_baseline}
\end{equation}
Modules yielding the lowest $I_{\text{zero}}$ scores are permanently masked out to satisfy the target compute reduction budget $\rho$. Collectively, these baselines represent diverse approaches for reducing transformer computation through early exiting, dynamic routing, learned gating, and structural pruning.
\raggedbottom 
\paragraph{Datasets.} Evaluation is performed on both language modeling and commonsense reasoning benchmarks. For language modeling, we report perplexity on WikiText-2, the Penn Treebank (PTB), and the C4 corpus. For downstream reasoning, we evaluate on HellaSwag, PIQA, CommonsenseQA (CSQA), and WinoGrande using standard multiple-choice accuracy. These benchmarks collectively assess the ability of compute-reduction methods to preserve language modeling performance, commonsense reasoning, and multiple-choice question answering accuracy under reduced computation budgets.

\subsection{Main Results}

Table~\ref{tab:main_results_all} compare CausalGate against recent inference optimization baselines under  5\%, 10\%, 20\%, 30\% and 40\% module-removal budgets, respectively. Performance is evaluated across language modeling tasks (WikiText-2 and C4 perplexity) and downstream commonsense reasoning benchmarks (HellaSwag, PIQA, CommonsenseQA, and WinoGrande). To characterize model robustness under progressive compute reduction, Figure~\ref{fig:ptb_degradation} reports the relative perplexity degradation on PTB across increasing budgets. 

To assess practical inference efficiency, Table~\ref{tab:efficiency} reports end-to-end latency, throughput, and hardware speedups obtained by applying the learned module rankings. These measurements confirm that theoretical compute savings translate into measurable runtime improvements on physical hardware. Finally, to evaluate scalability beyond TinyLlama-1.1B, we apply CausalGate to Qwen2.5-3B-Instruct and Llama-3.1-8B-Instruct. Figure~\ref{fig:scaling_large_models} summarizes the resulting WikiText-2 and C4 perplexities across module-removal budgets ranging from 0\% to 20\%, demonstrating the efficacy of our intervention-guided module ranking on larger transformer backbones.

\begin{table}[t]
\centering
\caption{Inference efficiency of CausalGate under increasing module-removal budgets.}
\label{tab:efficiency}
\resizebox{\columnwidth}{!}{%
\begin{tabular}{lcccc}
\toprule
\textbf{Model} & \textbf{Removal} & \textbf{Latency (s)} & \textbf{Tok/s} & \textbf{Speedup} \\
\midrule
TinyLlama-1.1B & 0\%  & 2.100 & 60.9 & 1.00$\times$ \\
TinyLlama-1.1B & 10\% & 1.954 & 65.5 & 1.08$\times$ \\
TinyLlama-1.1B & 20\% & 1.747 & 73.3 & 1.20$\times$ \\
\midrule
Qwen2.5-3B     & 0\%  & 3.980 & 32.2 & 1.00$\times$ \\
Qwen2.5-3B     & 10\% & 3.641 & 35.2 & 1.09$\times$ \\
Qwen2.5-3B     & 20\% & 3.393 & 37.7 & 1.17$\times$ \\
\bottomrule
\end{tabular}%
}
\end{table}

\section{Discussion}
 \label{sec:discussion}
\paragraph{Non-Uniform Causal Importance.}
The intervention-derived causal importance scores are highly non-uniform across transformer modules. While a small subset of modules induces substantial changes in the output distribution when intervened upon, the majority produce only negligible effects. In particular, several MLP submodules exhibit disproportionately large KL divergence values, indicating a dominant influence on the model's predictions, whereas many attention submodules, especially in later layers, contribute comparatively little. This concentration of causal importance suggests that transformer computation is inherently redundant and motivates selectively preserving highly influential modules during inference. A visualization of the learned causal importance distribution is provided in the Appendix (Figure~\ref{fig:causal_importance_heatmap}).

\paragraph{Recovery of the Causal Hierarchy.}
A key objective of CausalGate is to distill expensive intervention-derived importance estimates into a compact set of learned gate parameters. Figure~\ref{fig:gate_rank_heatmap} demonstrates that the learned gates recover much of the causal hierarchy identified through intervention analysis. The resulting gate rankings exhibit a Spearman correlation of 0.78 with the intervention-derived KL rankings, indicating strong agreement in the relative ordering of module importance. Although the Pearson correlation is lower (0.51), this behavior is expected since the ranking objective explicitly encourages ordering preservation rather than exact value matching. Furthermore, substantial overlap is observed among the highest- and lowest-ranked modules, suggesting that the learned gates successfully identify both critical and redundant components of the network. These results indicate that intervention-derived causal structure can be effectively distilled into lightweight gate parameters, eliminating the need for costly intervention analysis during inference.

\begin{figure}[t]
    \centering
    \includegraphics[width=\columnwidth]{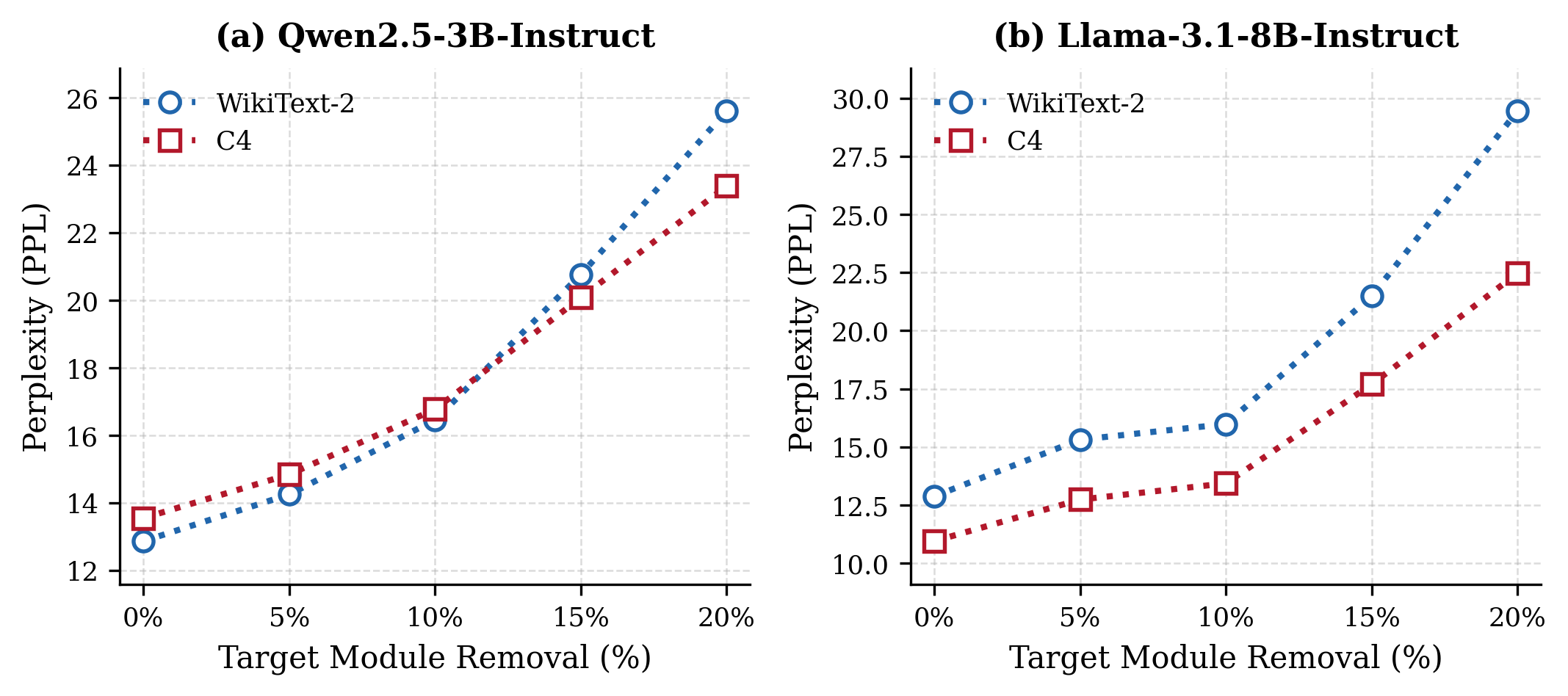}
    \caption{
    Scaling behavior of CausalGate on Qwen2.5-3B and Llama-3.1-8B under increasing module-removal budgets.}
    \label{fig:scaling_large_models}
\end{figure}

\begin{figure}[t]
\centering
\includegraphics[width=\columnwidth]{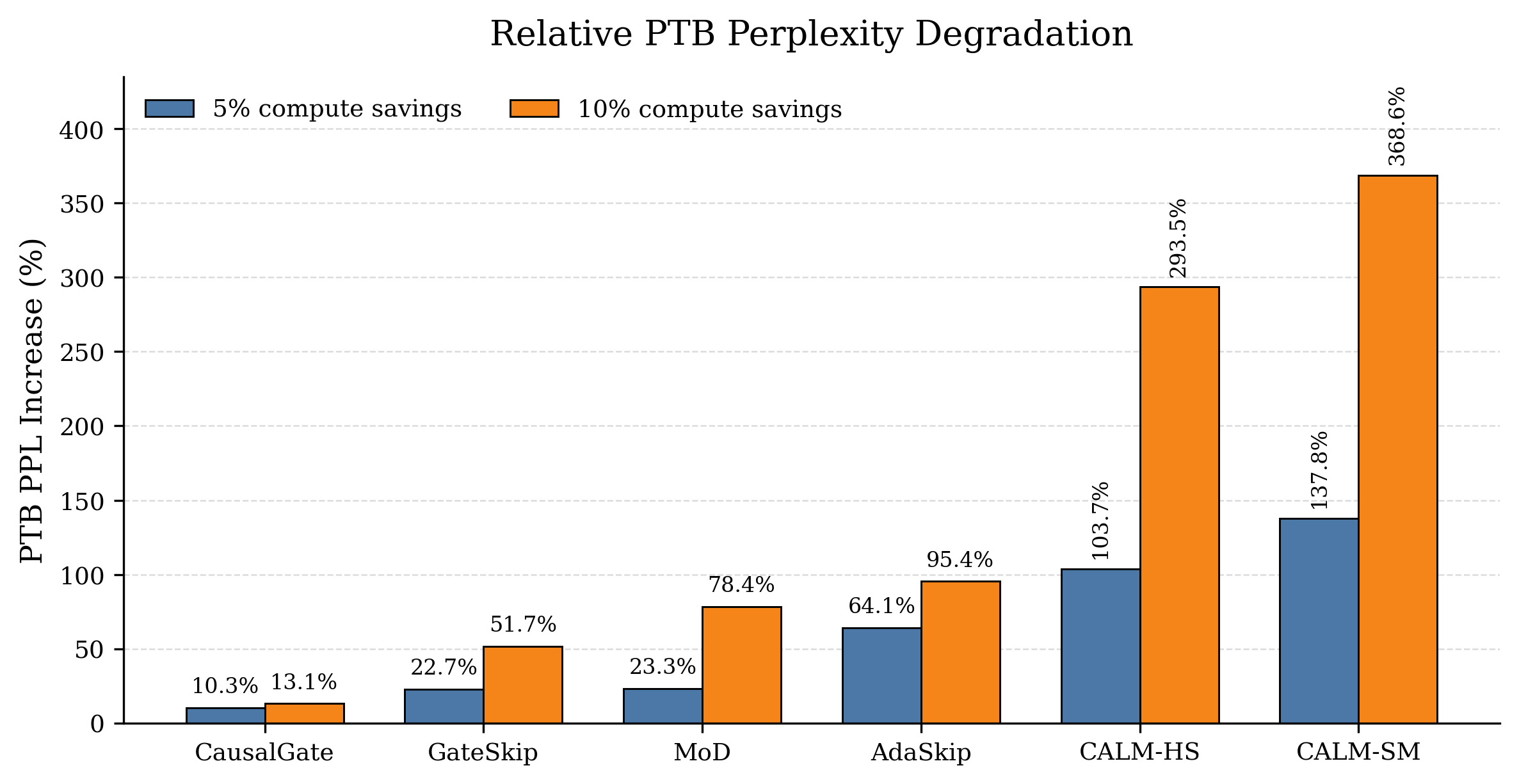}
\caption{
Relative perplexity degradation on the PTB benchmark under approximately 5\% and 10\% module-removal budgets. Values are reported as the percentage increase in perplexity relative to the full TinyLlama-1.1B model.
}
\label{fig:ptb_degradation}
\end{figure}

\paragraph{Impact on Compute Reduction.}
Table~\ref{tab:main_results_all} summarizes performance across progressive module-removal budgets from 5\% to 40\%. At the 5\% and 10\% tiers, CausalGate establishes a highly favorable Pareto-efficiency frontier, achieving the lowest C4 perplexities (10.84 and 12.84) alongside superior or highly competitive downstream reasoning accuracy compared to all baselines. This margin expands significantly in the moderate 20\% and 30\% compression regimes. Notably, at a 30\% budget reduction, CausalGate retains a tight language modeling perplexity (WikiText PPL of 116.0), whereas key baselines like GateSkip (365.0) and MoD (1313.0) experience catastrophic architectural degradation. Furthermore, the complete failure of the data-dependent Act-Norm baseline at these moderate budgets (perplexity exceeding 3500) empirically validates that active structural interventions are strictly necessary to diagnose true functional utility, which simple static observation misses. At the extreme 40\% threshold, all models encounter an architectural inflection point where removing nearly half the sub-layers severely drops performance. While CALM (Hidden-State) demonstrates localized resilience on language modeling at this edge, CausalGate maintains highly competitive downstream reasoning metrics. This systemic robustness is verified in Figure~\ref{fig:ptb_degradation}, where CausalGate incurs the lowest relative perplexity degradation on PTB across multiple compression tiers. Beyond single-model benchmarks, Figure~\ref{fig:scaling_large_models} shows that this intervention-guided scaling behavior successfully translates to Qwen2.5-3B and LLaMA-3.1-8B under progressive constraints. Crucially, these advantages incur no operational overhead; as detailed in Table~\ref{tab:efficiency}, compile-time mask generation translates theoretical structural savings into concrete throughput gains and latency reductions. A two-tailed Wilcoxon signed-rank test comparing the task-wise profiles of CausalGate against all baselines across the 5\%--40\% budgets confirms these performance gains are statistically significant ($p < 0.05$). Overall, these results confirm that intervention-derived causal importance provides a robust criterion for structural pruning.
\paragraph{Limitations and Future Work.}
The proposed framework employs a global module ranking that is learned once and applied uniformly across all inputs. While this design enables efficient module selection and stable compute reduction, it does not account for input-specific variability in module importance. Consequently, modules that are unimportant on average may still be beneficial for particular inputs or tasks. Future work could extend CausalGate toward input-adaptive policies that dynamically select modules based on the characteristics of each input sequence. Although the observed trends are consistent across both language modeling and reasoning benchmarks, further validation on larger language models and diverse architectures is needed to assess the generality of intervention-derived module ranking. Future work may also investigate finer-grained causal gating strategies, including token-level and context-dependent module selection, as well as integration with runtime-aware execution frameworks for achieving practical inference acceleration.
\section{Conclusion}
\label{sec:conclusion}
In this work, we introduced CausalGate, an intervention-guided framework for compute-efficient transformer inference. Unlike existing approaches that rely on observational heuristics such as confidence estimates or representation similarity, CausalGate directly measures module importance through structural interventions and quantifies each module’s causal influence on the model’s output distribution. By distilling intervention-derived importance scores into lightweight scalar gates using EMA target aggregation and pairwise ranking supervision, the proposed method learns a stable hierarchy of module importance that enables efficient module selection without requiring intervention analysis at inference time. Experimental results on language modeling and commonsense reasoning benchmarks demonstrate that CausalGate consistently achieves a more favorable compute-performance trade-off than existing adaptive inference methods, particularly under increasingly restrictive compute budgets. These findings suggest that causal contribution provides a principled criterion for identifying redundant computation in LLMs and offer a promising direction toward more efficient and interpretable inference systems.
\setlength{\bibsep}{1.5pt}
\bibliography{aaai2027}

\input{appendix}
\end{document}

%% file: appendix.tex
\clearpage 

\appendix
\normalsize

\section{Ablation Studies and Analysis}
\label{sec:ablations}
\begin{table}[H]

\centering
\caption{One-factor-at-a-time hyperparameter sensitivity analysis of CausalGate. Each hyperparameter is varied independently while all remaining settings are fixed to their default values. Spearman correlation is computed with respect to the intervention-derived ranking, and WikiText-2 perplexity is reported at the 10\% module-removal budget.}
\label{tab:hp_sensitivity}
\resizebox{\columnwidth}{!}{
\begin{tabular}{llccc}
\toprule
\textbf{Hyperparameter} & \textbf{Value} &
\textbf{Spearman}$\uparrow$ &
\textbf{WikiText PPL}$\downarrow$ &
\textbf{Selected} \\
\midrule
Learning Rate & 0.001 & 0.6879 & 20.19 & \\
              & 0.005 & 0.6765 & 21.15 & \\
              & \textbf{0.010} & 0.6849 & 18.46 & \checkmark \\
              & 0.020 & 0.681 & 18.52 & \\
\midrule
EMA $\beta$ & 0.70 & 0.679 & 18.6 & \\
            & 0.80 & 0.675 & 18.62 & \\
            & \textbf{0.90} & 0.68 & 18.44 & \checkmark \\
            & 0.95 & 0.66 & 19.01 & \\
\midrule
Target Floor $\alpha$ & 0.10 & 0.72 & 18.47 & \\
                      & \textbf{0.25} & 0.71 & 18.44  & \checkmark \\
                      & 0.40 & 0.7 & 18.44 & \\
\midrule
Rank Margin $\gamma$ & 0.01 & 0.62 & 18.48 & \\
                     & \textbf{0.05} & 0.68 & 18.43 & \checkmark \\
                     & 0.10 & 0.65 & 18.45 & \\
\midrule
Rank Pairs & 32 & 0.68 & 18.45 & \\
           & 64 & 0.68 & 18.45 & \\
           & \textbf{128} & 0.68 & 18.45 & \checkmark \\
           & 256 & 0.68 & 18.45 & \\
\midrule
$\lambda_{\mathrm{causal}}$ & 1 & 0.54 & 20.1 & \\
                            & 5 & 0.56 & 19.7 & \\
                            & \textbf{10} & 0.6 & 18.48 & \checkmark \\
                            & 20 & 0.6 & 18.9 & \\
\midrule
$\lambda_{\mathrm{rank}}$ & 0 & 0.43 & 19.2 & \\
                          & 1 & 0.57 & 19.2 & \\
                          & \textbf{2} & 0.68 & 18.46 & \checkmark \\
                          & 3 & 0.66 & 18.56 & \\
\midrule
Training Steps & 500 & 0.7 & 21.15 & \\
               & \textbf{1000} & 0.68 & 18.4 & \checkmark \\
               & 1500 & 0.69 & 18.45 & \\
\bottomrule
\end{tabular}
}
\end{table}

\section{Algorithm}
\input{algorithm1}
\input{algorithm2}

\section{An Interventional Causal Interpretation of CausalGate}
\label{app:causal}

Although CausalGate is implemented entirely within a transformer, its
module-ranking strategy can be interpreted through the lens of
interventional causal reasoning. Rather than estimating module importance
from observational statistics such as activation magnitudes, hidden-state
similarity, or confidence measures, CausalGate explicitly intervenes on
individual transformer modules and measures the resulting change in the
model prediction.





 \begin{figure}[H]
\centering
\includegraphics[width=0.9\columnwidth]{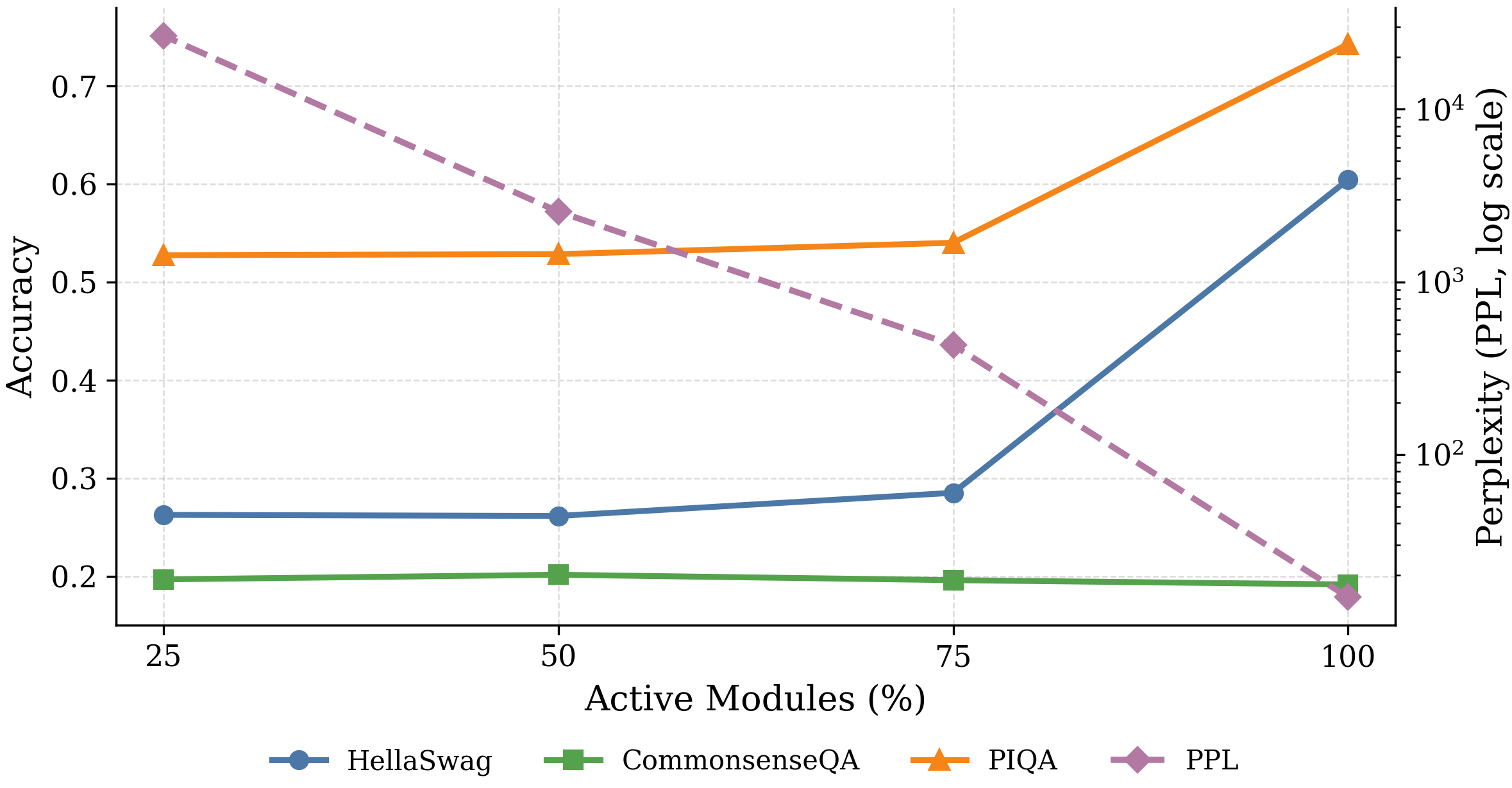}
\caption{
Effect of reducing the fraction of active transformer modules. Accuracy is reported for HellaSwag, CommonsenseQA, and PIQA, while perplexity is shown on a logarithmic scale.}
\label{fig:active_module_ablation}
\end{figure}

Figure~\ref{fig:active_module_ablation} shows the effect of reducing the fraction of active transformer modules on model performance. The figure compares accuracy changes on three benchmark tasks HellaSwag, CommonsenseQA, and PIQA along with changes in perplexity. By progressively decreasing the number of active modules, the figure demonstrates how model efficiency and performance are affected, highlighting the trade-off between reducing computational components and maintaining accuracy and language modeling quality.


Consider a transformer consisting of $N$ computational modules
(attention and MLP sublayers). Let
$X$ denote the input token sequence,
$M_i$ denote the output representation of the $i$-th transformer module,
and $Y$ denote the next-token predictive distribution.

\begin{figure}[H]
    \centering
    \includegraphics[width=0.50\textwidth]{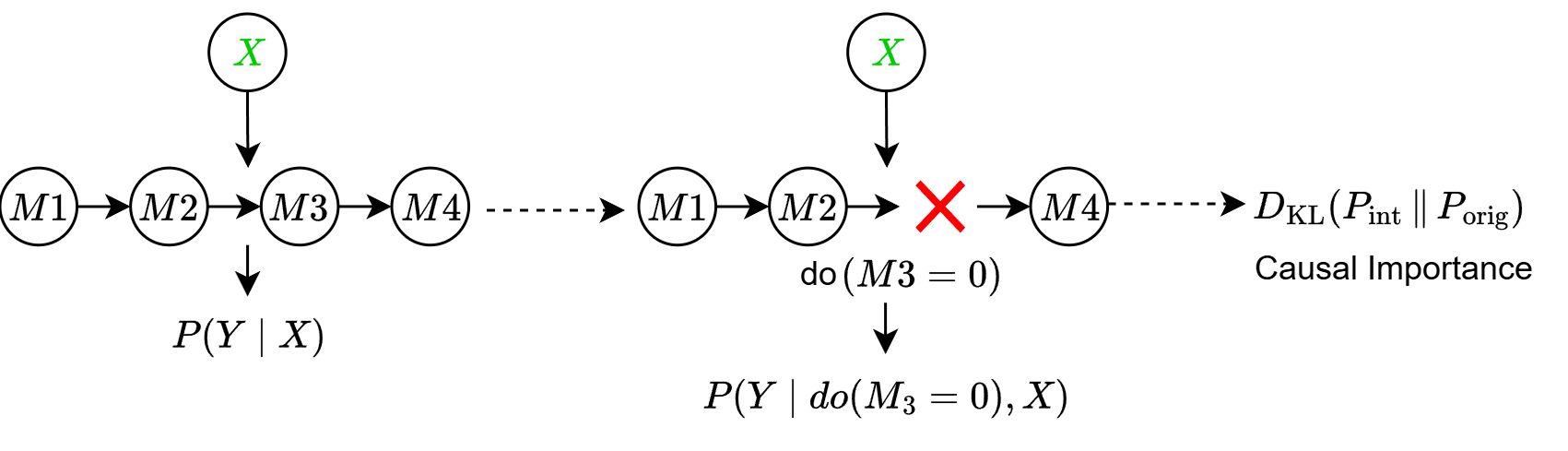}
    \caption{
    Interventional causal interpretation of CausalGate. Given an input sequence $X$, the transformer produces the predictive distribution $P(Y\mid X)$. An intervention suppresses the residual contribution of a selected module (illustrated for $M_3$), producing the intervened distribution $P(Y\mid do(M_3=0),X)$. The KL divergence between the original and intervened distributions quantifies the module's interventional influence on the model prediction, providing the supervision signal for learning module importance.
    }
    \label{fig:causal_intervention}
\end{figure}

The forward computation may be written as

\begin{align}
M_1 &= f_1(X),\\
M_2 &= f_2(M_1),\\
&\vdots\\
M_N &= f_N(M_{N-1}),\\
P(Y|X) &= g(M_N),
\end{align}

where each module causally influences all downstream representations
through the residual computation graph. Under this interpretation, each
transformer module represents an intermediate causal variable whose
effect propagates to the final prediction.
\\
Figure~\ref{fig:causal_intervention} illustrates this computational
graph together with the intervention performed by CausalGate. Rather than estimating module importance from observational statistics, an intervention suppresses the residual contribution of a selected transformer module while leaving the remainder of the computational graph unchanged. This produces an intervened predictive distribution that can be directly compared with the original model prediction. From an interventional perspective, modules whose removal induces larger changes in the output distribution are interpreted as exerting greater causal influence on the model's prediction. Consequently, CausalGate estimates module importance through explicit interventions on the computational graph rather than through correlation-based proxies such as activation magnitude or hidden-state similarity.



\begin{table}[H]

\centering
\caption{Summary of datasets used for training and evaluation.}
\label{tab:dataset_metadata}
\resizebox{\columnwidth}{!}{
\begin{tabular}{lccc}
\toprule
\textbf{Dataset} & \textbf{Usage} & \textbf{Task} & \textbf{License} \\
\midrule
WikiText-2 & Training + Evaluation & Language Modeling & CC BY-SA 3.0 \\
C4 & Evaluation & Language Modeling & ODC-BY 1.0 \\
HellaSwag & Evaluation & Commonsense Reasoning & MIT \\
PIQA & Evaluation & Physical Reasoning & MIT \\
CommonsenseQA & Evaluation & Question Answering & MIT \\
WinoGrande & Evaluation & Commonsense Reasoning & MIT \\
\bottomrule
\end{tabular}
}
\end{table}

Table~\ref{tab:dataset_metadata} represents the summary of datasets used for training and evaluation. WikiText-2 is the only dataset used for both training and evaluation, supporting language modeling, while C4 is used solely for evaluating language modeling performance. The other datasets HellaSwag, PIQA, CommonsenseQA, and WinoGrande are used exclusively for evaluation, covering commonsense reasoning, physical reasoning, and question answering tasks. 

\begin{figure}[H]
\centering

\includegraphics[width=\textwidth]{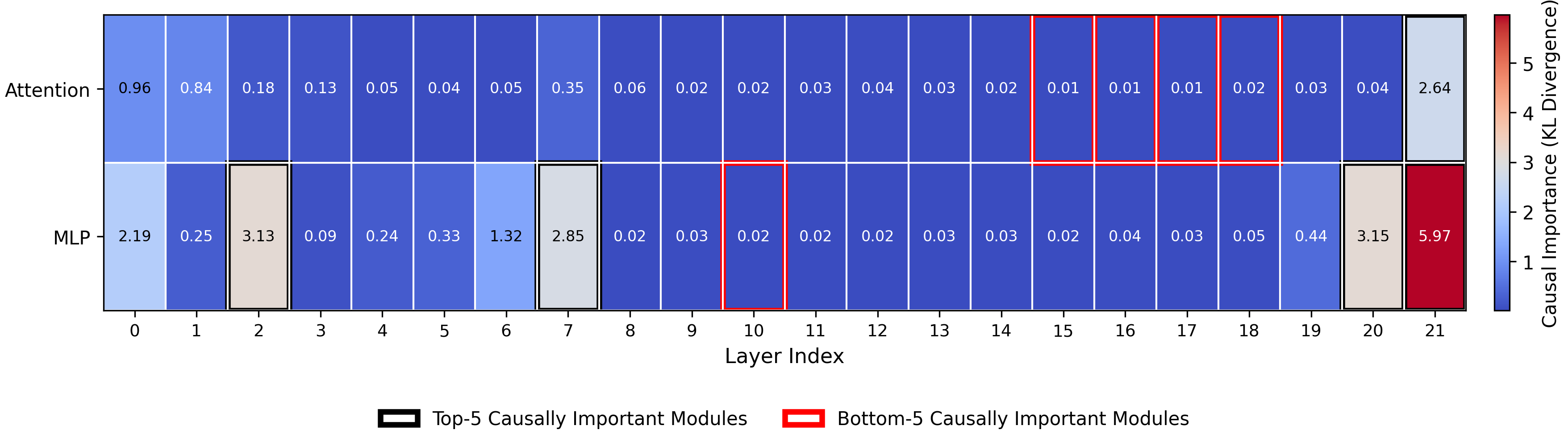}
\caption{
Module-level causal importance scores obtained through intervention analysis. Each cell reports the average KL divergence induced by zeroing the corresponding attention or MLP submodule. Higher values indicate modules whose removal causes larger changes in the model's output distribution.
}
\label{fig:causal_importance_heatmap}
\end{figure}

Figure~\ref{fig:causal_importance_heatmap} illustrates the module-level causal importance scores obtained through intervention analysis. The figure shows the impact of removing individual attention and MLP submodules by measuring the resulting change in the model’s output distribution using average KL divergence. Higher KL divergence values indicate modules that have a greater influence on the model’s behavior, as their removal leads to larger deviations in the output. This visualization highlights the relative importance of different model components and identifies the submodules that contribute most significantly to the model’s predictions. 
\onecolumn

%% file: algorithm1.tex



\begin{algorithm}[H]
\large
\caption{Intervention-Based Causal Importance Estimation}
\label{alg:causal_scan}

\begin{algorithmic}[1]
\Require Transformer $f$, module set $\mathcal{M}=\{m_i\}_{i=1}^{M}$, input batch $x$
\Ensure Module-level causal scores $\{\Delta_i\}_{i=1}^{M}$

\State Run the original model to obtain logits $\ell_{\mathrm{orig}} \leftarrow f(x)$
\State Compute final-token distribution $p_{\mathrm{orig}}(y_T|x)$

\For{each module $m_i \in \mathcal{M}$}
    \State Zero the residual-branch output of module $m_i$
    \State Run the intervened model to obtain logits $\ell_{\mathrm{int},i}$
    \State Compute final-token distribution $p_{\mathrm{int},i}(y_T|x)$
    \State Compute causal importance score:
    \[
    \Delta_i =
    D_{\mathrm{KL}}
    \left(
    p_{\mathrm{int},i}(y_T|x)
    \;\|\;
    p_{\mathrm{orig}}(y_T|x)
    \right)
    \]
\EndFor

\State \Return $\{\Delta_i\}_{i=1}^{M}$

\end{algorithmic}
\end{algorithm}

%% file: algorithm2.tex
\begin{algorithm}[H]
\caption{CausalGate Distillation and Module Selection}
\label{alg:causalgate_distillation}
\begin{algorithmic}[1]
\Require Frozen transformer $f$, training data $\mathcal{D}$, module set $\mathcal{M}$
\Require Target floor $\alpha$, EMA decay $\beta$, ranking margin $\gamma$, ranking pairs $R$
\Require Loss weights $\lambda_{\mathrm{LM}}, \lambda_{\mathrm{sparse}}, \lambda_{\mathrm{causal}}, \lambda_{\mathrm{rank}}$
\Ensure Learned gate ranking and selected modules under target budget $\rho$

\State Initialize one scalar gate logit $\theta_i \leftarrow 0$ for each module $m_i$
\State Set $g_i \leftarrow \sigma(\theta_i)$
\State Freeze all backbone parameters of $f$

\For{training step $t=1,\ldots,T$}
    \State Sample batch $x \sim \mathcal{D}$
    \State Compute intervention scores $\{\Delta_i^{(t)}\}_{i=1}^{M}$ using Algorithm~\ref{alg:causal_scan}

    \State Normalize scores:
    \[
    \tilde{z}_i^{(t)}
    =
    \frac{\Delta_i^{(t)}}{\max_j \Delta_j^{(t)}+\epsilon}
    \]

    \State Apply target floor:
    \[
    z_i^{(t)}
    =
    \alpha + (1-\alpha)\tilde{z}_i^{(t)}
    \]

    \If{$t=1$}
        \State Initialize EMA target $\bar{z}_i^{(t)} \leftarrow z_i^{(t)}$
    \Else
        \State Update EMA target:
        \[
        \bar{z}_i^{(t)}
        =
        \beta \bar{z}_i^{(t-1)}
        +
        (1-\beta)z_i^{(t)}
        \]
    \EndIf

    \State Update gate values $g_i \leftarrow \sigma(\theta_i)$

    \State Compute causal regression loss:
    \[
    \mathcal{L}_{\mathrm{MSE}}
    =
    \frac{1}{M}
    \sum_{i=1}^{M}
    \left(g_i-\bar{z}_i^{(t)}\right)^2
    \]

    \State Sample $R$ random non-tied module pairs $\mathcal{P}$
    \State Define $s_{ij}=\operatorname{sign}(\bar{z}_i^{(t)}-\bar{z}_j^{(t)})$

    \State Compute ranking loss:
    \[
    \mathcal{L}_{\mathrm{rank}}
    =
    \frac{1}{|\mathcal{P}|}
    \sum_{(i,j)\in\mathcal{P}}
    \max
    \left(
    0,\,
    \gamma-s_{ij}(g_i-g_j)
    \right)
    \]

    \State Compute sparsity loss:
    \[
    \mathcal{L}_{\mathrm{sparse}}
    =
    \frac{1}{M}
    \sum_{i=1}^{M}g_i^2
    \]

    \State Compute total loss:
    \[
    \mathcal{L}
    =
    \lambda_{\mathrm{LM}}\mathcal{L}_{\mathrm{LM}}
    +
    \lambda_{\mathrm{sparse}}\mathcal{L}_{\mathrm{sparse}}
    +
    \lambda_{\mathrm{causal}}\mathcal{L}_{\mathrm{MSE}}
    +
    \lambda_{\mathrm{rank}}\mathcal{L}_{\mathrm{rank}}
    \]

    \State Update only gate logits $\{\theta_i\}_{i=1}^{M}$ using AdamW
\EndFor

\State Rank modules by learned gate values in descending order
\State For target saved ratio $\rho$, compute $S=\operatorname{round}(\rho M)$ and $K=M-S$
\State Keep the top-$K$ modules and suppress the remaining $S$ modules
\State \Return learned gate ranking and selected module set
\end{algorithmic}
\end{algorithm}